# A topological insight into restricted Boltzmann machines

**Decebal Constantin Mocanu · Elena Mocanu · Phuong H. Nguyen · Madeleine Gibescu · Antonio Liotta**

**Pre-print version.** *Published as a journal article in Machine Learning, ECML PKDD 2016 special issue, issn 1573-0565, doi 10.1007/s10994-016-5570-z.*

**Abstract** Restricted Boltzmann Machines (RBMs) and models derived from them have been successfully used as basic building blocks in deep artificial neural networks for automatic features extraction, unsupervised weights initialization, but also as density estimators. Thus, their generative and discriminative capabilities, but also their computational time are instrumental to a wide range of applications. Our main contribution is to look at RBMs from a topological perspective, bringing insights from network science. Firstly, here we show that RBMs and Gaussian RBMs (GRBMs) are bipartite graphs which naturally have a small-world topology. Secondly, we demonstrate both on synthetic and real-world datasets that by constraining RBMs and GRBMs to a scale-free topology (while still considering local neighborhoods and data distribution), we reduce the number of weights that need to be computed by a few orders of magnitude, at virtually no loss in generative performance. Thirdly, we show that, for a fixed number of weights, our proposed sparse models (which by design have a higher number of hidden neurons) achieve better generative capabilities than stan-

Decebal Constantin Mocanu
Department of Electrical Engineering, Eindhoven University of Technology, the Netherlands
Tel.: +31 40-247 5394
E-mail: d.c.mocanu@tue.nl

Elena Mocanu
Department of Electrical Engineering, Eindhoven University of Technology, the Netherlands
Tel.: +31 40-247 8464
E-mail: e.mocanu@tue.nl

Phuong H. Nguyen
Department of Electrical Engineering, Eindhoven University of Technology, the Netherlands
+31 40-247 5830
E-mail: p.nguyen.hong@tue.nl

Madeleine Gibescu
Department of Electrical Engineering, Eindhoven University of Technology, the Netherlands
+31 40-247 8258
E-mail: m.gibescu@tue.nl

Antonio Liotta
Department of Electrical Engineering, Eindhoven University of Technology, the Netherlands
+31 40-247 8311
E-mail: a.liotta@tue.nl

dard fully connected RBMs and GRBMs (which by design have a smaller number of hidden neurons), at no additional computational costs.

## 1 Introduction

Since its conception, deep learning (Bengio, 2009) is widely studied and applied, from pure academic research to large-scale industrial applications, due to its success in different real-world machine learning problems such as audio recognition (Lee et al, 2009), reinforcement learning (Mnih et al, 2015), transfer learning (Ammar et al, 2013), and activity recognition (Mocanu et al, 2015). Deep learning models are artificial neural networks with multiple layers of hidden neurons, which have connections only among neurons belonging to consecutive layers, but have no connections within the same layers. In general, these models are composed by basic building blocks, such as Restricted Boltzmann Machines (RBMs) (Smolensky, 1987). In turn, RBMs have proven to be successfully not just providing good initialization weights in deep architectures (in both supervised and unsupervised learning), but also as standalone models in other types of applications. Examples are density estimation to model human choice (Osogami and Otsuka, 2014), collaborative filtering (Salakhutdinov et al, 2007), information retrieval (Gehler et al, 2006), or multi-class classification (Larochelle and Bengio, 2008). Thus, an important research direction is to improve the performance of RBMs on any component (e.g. computational time, generative and discriminative capabilities).

The main contribution of this paper is to look at the deep learning basic building blocks, i.e. RBMs and Gaussian RBMs (GRBMs) (Hinton and Salakhutdinov, 2006), from a topological perspective, bringing insights from network science, an extension of graph theory which analyzes real world complex networks (Strogatz, 2001). Firstly, we study the topological characteristics of RBMs and GRBMs, finding that these exhibit a small-world topology (Watts and Strogatz, 1998). We then hypothesize that by constraining the topology to be also scale-free (Barabasi and Albert, 1999) it is possible to reduce the size of ordinary RBMs and GRBMs models, as it has been shown by Del Genio et al (2011) that scale-free networks are sparse. We introduce a method to make small-world, scale-free topologies while still considering local neighborhoods and data distribution. We dub the resulting models as compleX Boltzmann Machine (XBM) and Gaussian compleX Boltzmann Machine (GXBM), respectively. An interesting finding is that constraining such XBM and GXBM topologies at their inception leads to intrinsically sparse networks, a considerable advantage to typical state-of-the-art methods in which sparsity is enforced as an aftermath, that is during testing (exploitation) phase (Swersky et al, 2012; Wan et al, 2015; Ranzato et al, 2008; Luo et al, 2011; Lee et al, 2008). In turn, XBM and GXBM have a considerably smaller number of weights, which further on contributes to considerably faster computational times (proportional to the number of weights in the model), both in the training and testing phases. What is more, we found that the proposed topology imposes an inductive bias on XBMs and GXBMs, which leads to better statistical performance than RBMs and GRBMs. Our comparative study is based on both simulated and real-world data, including the Geographical origin of music dataset (Zhou et al, 2014), the MNIST digits dataset, CalTech 101 Silhouettes dataset (Marlin et al, 2010), and the 8 datasets from UCI evaluation suite (Larochelle and Murray, 2011). We show that, given the same number of hidden neurons, XBM and GXBM have similar or relatively close capabilities to RBM and GRBM, but are considerably faster thanks to their reduced amount of weights. For instance, in a network of 100 visible and 100 hidden neurons, the reduction in weights was by one order of magnitude. A network with 1000 visible and 1000 hidden neurons led to a reduction in weights by two orders of magnitude. Additionally, we show that given the same amount of weights, RBMs or GRBMs derived models with a higher number of hidden neurons and sparse connectivity achieve better generative capabilities than fully connected RBMs or GRBMs with a smaller amount of hidden neurons.

The remaining of this paper is organized as follows. Section 2 presents background knowledge about Boltzmann machines and complex networks for the benefit of the non-specialist reader and highlights the key motivations of our work. Section 3 discusses the relation between deep leaning and network science and details the mathematical models of our proposed methods. Section 4 describes the experiments performed and analyzes the results. Finally, Section 5 concludes the paper and presents directions of future research.

## 2 Background and motivations

### 2.1 Boltzmann machines

Originally derived by Ackley et al (1985), a Boltzmann machine is a network of symmetrically connected stochastic binary units (or neurons). To formalize a Boltzmann machine, and its variants, three main ingredients are required, namely an energy function providing scalar values for a given configuration of the network, the probabilistic inference and the learning rules required for fitting the free parameters. This bidirectional connected network with stochastic nodes has no unit connected with itself. However, Boltzmann machines with unconstrained connectivity are infeasible in practical problems due to the intractable inference. A critical step in taming computational complexity is to add constraints to the connectivity network, which is what makes Boltzmann machines applicable to real world problems.

Smolensky (1987) presented restricted Boltzmann machine that could learn a probability distribution over its set of inputs. The model architecture was restricted by not allowing intra-layer connections between the units, as depicted in Figure 2 (left). Since their conception, different types of Boltzmann machines have been developed and successfully applied. Yet most of these variations preserve some fundamental characteristics. RBMs are generative stochastic neural networks which consists of two binary layers, the visible layer, $\mathbf{v} = [v_1, v_2, .., v_{n_v}]$, and the hidden layer, $\mathbf{h} = [h_1, h_2, .., h_{n_h}]$, with $n_v$ being the number of visible neurons and $n_h$ the number of the hidden ones. Formally, the energy function of RBMs for any state $\{\mathbf{v}, \mathbf{h}\}$ is computed by summing over all possible interactions between neurons, weights and biases, as follows:

$$E(v, h) = -\sum_{i,j} v_i h_j W_{ij} - \sum_i v_i a_i - \sum_j h_j b_j \quad (1)$$

where $W_{ij}$ denotes the connection between the visible neuron $i$ and the hidden neuron $j$, $a_i$ is the bias for visible neuron $i$ and $b_j$ is the bias for hidden neuron $j$. The term $\sum_{i,j} v_i h_j W_{ij}$ represents the total energy between neurons from different layers, $\sum_i v_i a_i$ represents the energy of the visible layer and $\sum_j h_j b_j$ the energy of the hidden layer. The inference in RBMs is stochastic. For any hidden neuron $j$ the conditional probability is given by $p(h_j|\mathbf{v}) = \mathcal{S}(b_j + \sum_i v_i W_{ij})$, and for any visible unit $i$ it is given by $p(v_i|\mathbf{h}) = \mathcal{S}(a_i + \sum_j h_j W_{ij})$, where $\mathcal{S}(\cdot)$ is a sigmoid function.

Later on, Hinton and Salakhutdinov (2006) have extended the RBMs models to make them suitable for a large number of applications with real-valued feature vectors. They used exponential family harmoniums results from Welling et al (2005) and developed the Gaussian Restricted Boltzmann Machine (GRBM) model that, like RBMs, forms a symmetrical bipartite graph. However, the binary units from the visible layer $\mathbf{v}$ are replaced by linear units with Gaussian noise. The hidden units $\mathbf{h}$ remain binary. Therein, the total energy function for a state $\{\mathbf{v}, \mathbf{h}\}$ of GRBMs is calculated in a similar manner to RBMs, but includes a slight change to take into consideration the Gaussian noise of the visible neurons, as defined in Equation 2.

$$E(v,h) = -\sum_{i,j} \frac{v_i}{\sigma_i} h_j W_{ij} - \sum_i \frac{(v_i - a_i)^2}{2\sigma_i^2} - \sum_j h_j b_j \tag{2}$$

where, the term $\sum_{i,j} \frac{v_i}{\sigma_i} h_j W_{ij}$ gives the total energy between neurons from different layers, $\sum_i \frac{(v_i - a_i)^2}{2\sigma_i^2}$ is the energy of the visible layer, and $\sigma_i$ represents the standard deviation of the visible neuron $i$. The stochastic inference for any hidden neuron $j$ can be done as for RBMs, while for any visible unit $i$ is made by sampling from a Gaussian distribution, defined as $\mathcal{N}(a_i + \sum_j h_j W_{ij}, \sigma_i^2)$.

Parameters of RBM and GRBM models are fitted by maximizing the likelihood function. In order to maximize the likelihood of the model, the gradients of the energy function with respect to the weights have to be calculated. Because of the difficulty in computing the derivative of the log-likelihood gradients, Hinton (2002) proposed an approximation method called Contrastive Divergence (CD). In maximum likelihood, the learning phase actually minimizes the Kullback-Leiber (KL) measure between the input data distribution and the model approximation. Thus, in CD, learning follows the gradient of:

$$CD_n \propto D_{KL}(p_0(\mathbf{x})||p_\infty(\mathbf{x})) - D_{KL}(p_n(\mathbf{x})||p_\infty(\mathbf{x})) \tag{3}$$

where, $p_n(.)$ is the resulting distribution of a Markov chain running for $n$ steps. Besides that, other methods have been proposed to train RBMs (e.g. persistent contrastive divergence (Tieleman, 2008), fast persistent contrastive divergence (Tieleman and Hinton, 2009), parallel tempering (Desjardins et al, 2010)), or to replace the Gibbs sampling with a transition operator for a faster mixing rate and to improve the learning accuracy without affecting computational costs (Brgge et al, 2013).

### 2.2 Sparsity in restricted Boltzmann machines

In general and for the purposes of machine learning, obtaining a sparse version of a given model, leads to a reduction in parameters, which, in turns helps in addressing problems such as overfitting and excessive computational complexity. The sparsity issue in RBMs is so important that considerable attention is given to it in the literature. Hereafter we point to the most relevant works but will not attempt to provide a comprehensive overview. It is worth mentioning the work by Dieleman and Schrauwen (2012) and Yosinski and Lipson (2012) who have shown how the histogram of the RBM weights changes shape during the training process, going from a Gaussian shape (initially) to a shape that peaks around zero (which provides a further motivation towards sparsity enforcement).

One of the most common methods to obtain sparse representations is by encouraging it during the training phase using different variants of a sparsity penalty function, as done for instance, in Lee et al (2008) and Ranzato et al (2008). However, performing this process during the learning phase does not guarantee sparsity in the testing phase (Swersky et al, 2012). To overcome this limitation, Cardinality-RBM (Ca-RBM) was proposed by Swersky et al (2012) to ensure sparsity in the hidden representation by introducing cardinality potentials into the RBMs energy function. Moreover, Wan et al (2015) have proposed Gaussian Ca-RBM, in which they replace the universal threshold for hidden units activation from Ca-RBM with adaptable thresholds. These thresholds are sampled from a certain distribution which takes into consideration the input data. Recently, Han et al (2015) introduced one of the most efficient methods to obtain weights sparsity in deep neural network. They successfully obtained up to 10 times less weights in deep neural networks with no loss in accuracy. The method assumes three simple steps: (1) the network is trained to learn the most important connections; (2) the unimportant connections are pruned; (3) the network is retrained to fine tune the weights of the remaining connections. To achieve the best performance the steps 2 and 3 have to be repeated iteratively making it a computationally expensive method.

Thus, to the best of our knowledge, all of the state-of-the-art methods impose a sparsity regularization target during the learning process, which makes them impractical with large datasets having millions (or even billions) of input features. This is because the training process is excessively slow in such situations. Our solution to overcome this problem is to ensure weight sparsity from the initial design of an RBM using relevant findings from the field of network science. To this end, next section introduces these findings.

### 2.3 Complex networks

Complex networks (e.g. biological neural networks, actors and movies, power grids, transportation networks) are everywhere, in different forms and different fields, from neurobiology to statistical physics (Strogatz, 2001), and they are studied in network science. Formally, a complex network is a graph with non trivial topological features, human or nature made. The most two well-known and deeply studied types of topological features in complex networks are the scale-free and the small-world concepts, due to the fact that a wide range of real-world complex networks have these topologies. A network with a scale-free topology (Barabasi and Albert, 1999) is a sparse graph (Del Genio et al, 2011) that approximately has a power-law degree distribution $P(k) \sim k^{-\gamma}$, where the fraction $P(k)$ from the total nodes of the network has $k$ connections to other nodes, and the parameter $\gamma \in (2, 3)$ usually.

At the same time, a network model with the small-world topological feature (Watts and Strogatz, 1998) is defined to be a graph in which the typical distance ($L$) between two randomly chosen nodes (the number of hops required to reach one node from the other) is very small, approximately on the logarithmic scale with respect to the total number of nodes ($N$) in the network, while at the same time it is characterized by a clustering coefficient which is significantly higher than may appear by random chance. More formally, a graph sequence $(G_N)_{N \geq 1}$ has a small-world topology, if there is a constant $0 < K < \infty$ such that $lim_{N \to \infty} p(L_N \leq KlogN) = 1$, where $L_N$ is the typical shortest path of $G_N$ (van der Hofstad, 2016). As an example, Figure 1 roughly illustrates a small-world topology, and a scale-free one, in two small randomly generated graphs. Both types of topologies are studied below in the context of restricted Boltzmann machines, leading to our proposal of sparse Boltzmann machine models.

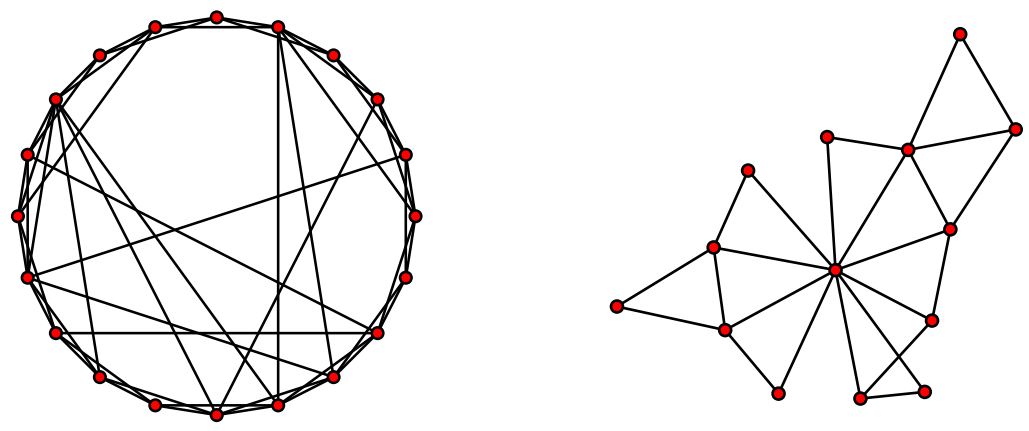

Fig. 1: Examples of complex networks topologies: (left) small-world; (right) scale-free.

## 3 Complex networks and Boltzmann machines

In this section, we firstly discuss the relation between complex networks on one side and restricted Boltzmann machines and Gaussian restricted Boltzmann machine on the other side. Secondly, we introduce an algorithm to generate sparse topologies for bipartite graphs which have both properties (i.e. scale-free and small-world) that also considers the distribution of the training data. Finally, we make use of the previous mentioned topology generator algorithm and present the mathematical details of two novel types of Boltzmann machines, dubbed compleX Boltzmann Machines (XBMs) and Gaussian compleX Boltzmann Machines (GXBMs).

### 3.1 Topological insight into RBMs and GRBMs

Lately, the neural networks of the human brain have started to be studied using tools from network science (Pessoa, 2014). It has been found that these exhibit both a small-world topology (i.e. the shortest path between any two nodes or neurons is very small, approximately equal to the logarithm of the total number of nodes) and a scale-free topology (i.e their degree distribution follows a power law). At the same time, by making small steps towards mimicking the architecture and the functionality of the brain, deep learning methods have emerged as a promising solution in computer science to develop automated learning systems (Jones, 2014; Mnih et al, 2015). Here we argue that there is a clear relation between network science and deep learning. In the scope of these arguments, we introduce the following proposition:

**Proposition 1** *Both restricted Boltzmann machines and Gaussian restricted Boltzmann machines are bipartite graphs which have a small-world topology.*

*Proof* The diameter (i.e. the longest shortest path between any two neurons) of RBMs or GRBMs is 2, independently on the number of hidden or visible neurons, due to the fact that both models have all the possible interlayer connections, but no intralayer connections. This yields that $L$ is bounded up by 2 for any RBM or GRBM. By replacing $L$ in the small-world definition from Subsection 2.3, we obtain $lim_{N \to \infty} p(2 \leq KlogN) = 1$, which is true for any constant $K$, $0 < K < \infty$[1]. Similarly as RBMs and GRBMs are complete bipartite graphs, their clustering coefficient (Latapy et al, 2008) is 1, being higher than any other possible cluster coefficient[2]. Thus, it is clear that any RBMs or GRBMs have a small-world topology.

[1] Please note that according to the definitions from van der Hofstad (2016), this reflects even a particular subset of small-worlds, namely ultra small-worlds.

[2] Please note that the clustering coefficient takes values between 0 and 1.

Following the same line, our intuition is that by preserving the small-world property of RBMs or GRBMs, while introducing also the scale-free property in their topology, we can obtain new sparse Boltzmann machines derived models which may have similar performance to RBMs or GRBMs, but with fewer free parameters (i.e. the weights between the visible and the hidden neurons). Thus, further on, we introduce the complex Boltzmann machine and the Gaussian complex Boltzmann machine (i.e. the derivatives of RBM and GRBM, respectively) which exhibit both scale-free and small-world topological properties.

### 3.2 Topology generation algorithm for XBM and GXBM

To generate a sparse topology in XBM and GXBM, we have devised a three stages heuristic method, detailed in Algorithm 1. In the first stage, a scale-free bipartite graph is generated; in the second one, the graph is adjusted to be also small-world; and in the third stage, the graph topology is fitted to the data distribution. Below, this method is thoroughly discussed.

First we generate a power-law degree sequence with $n_v + n_h$ elements, using $P(k) = k^{-\gamma}$, $\forall k \in \mathbb{N}$, $1 \leq k \leq n_v + n_h$ , with minimum degree equal to four to favor the small-world topology, and we sort it in descending order (i.e. Algorithm 1, lines 8-9). Each element from the sequence represents a node in the network and the actual value of that element (i.e. the degree) represents the number of connections for that specific node. After that, we split the degree sequence in two, by alternatively picking the nodes starting from those with the highest degree and proceeding until the smallest degree nodes are reached. In this way we populate two separate lists, for the hidden layer and for the visible layer, respectively (i.e. Algorithm 1, lines 10-16). Once the list having the smallest number of elements is completed, we add all the remaining elements of the original sequence to the bigger list (i.e. Algorithm 1, lines 17-22). In an undirected bipartite graph both layers need to have an equal number of connections. Due to the fact that the sum of the elements from one list might not be equal with the sum of the elements from the other list, we add some more degrees to the elements of the list with less degrees (proportionally to its initial degree distribution) to equalize the two lists (i.e. Algorithm 1, lines 23-28). Next, starting from these two lists, we create a bipartite graph $G$ using a Havel-Hakimi procedure (Hakimi, 1962) (i.e. Algorithm 1, line 29). Further on, we add few more connections to each node in $G$ with the aim to achieve the optimal clustering coefficient as required in small-world topologies (i.e. Algorithm 1, lines 30-49). This ensures also a dense local connectivity useful, by example, for images. To clarify how this is done, note that in Algorithm 1 (line 3[3]), when parameter $\sigma^{neigh}$ is increased the nodes local neighborhoods gradually turn into larger neighborhoods (or even the graph as a whole). In turn, when parameter $\phi$ is increased, the neighborhoods tend to become denser. The whole algorithm proceeds iteratively until the bipartite graph meets the criteria of small-worldness (i.e. Algorithm 1, line 51). We should mention, though, that during our whole study we observed that this property was usually achieved after just one iteration. To take the data distribution into consideration, as a final stage, the algorithm re-arranges the order of the visible nodes in $G$ such that the nodes having more connections end up corresponding to the training data features with a higher standard deviation (i.e. Algorithm 1, line 53). The resulting bipartite graph $G$ can then be used as the topology of our XBM or GXBM models (i.e. Algorithm 1, line 55-57), as detailed next.

[3] In this paper, we have varied $\sigma^{neigh}$ and $\phi$ between 4 and 6 to favor the emergence of local medium connected neighborhoods.

*%% define $n_v$, $n_h$ (number of visible and hidden neurons, respectively)*;
*%% assumptions: $n_v > 4$ and $n_h > 4$ (we consider non trivial cases)*;
*%% define $\sigma^{neigh}$, $\phi$ (parameters to control the nodes local connectivity)*;
*%% initialization*;
**set** $L^{th} = log(n_v + n_h)$;*%% set a threshold for the small-world topology*;
*%% topology generation*;
**repeat**
  generate randomly $S^{PL}$, a power law degree sequence of size $n_v + n_h$ with the minimum degree of 4;
  sort $S^{PL}$ in descending order;
  set $S^v$=[] and $S^h$=[]; *%% sequences to store the degree of the visible and hidden nodes,respectively*;
  $i$=1;
  **while** $i \leq 2 \times min(n_v, n_h)$ **do**
    $S^v$.append($S^{PL}[i]$);
    $S^h$.append($S^{PL}[i+1]$);
    i=i+2;
  **end**
  **if** $(n_v > n_h)$ **then**
    $S^v$.append($S^{PL}[2 \times n_h : end]$);
  **end**
  **else**
    $S^h$.append($S^{PL}[2 \times n_v : end]$);
  **end**
  **if** $sum(S^v) < sum(S^h)$ **then**
    add $sum(S^h) - sum(S^v)$ degrees equally distributed among the visible nodes;
  **end**
  **else**
    add $sum(S^v) - sum(S^h)$ degrees equally distributed among the hidden nodes;
  **end**
  $G$ =createBipartiteGraphUsingHavelHakimiProcedure($S^v$, $S^h$) (Hakimi, 1962);
  **for** *o=1:$\phi$* **do**
    **for** *i=1:$n_v$* **do**
      **while** *a finite number of trials* **do**
        $j = \lceil \mathcal{N}((i \times n_h)/n_v, \sigma^{neigh}) \rceil$;*%% sampled from a Gaussian distribution*;
        **if** $0 \leq j \leq n_h$ **then**
          addEdge $(i, j)$ to $G$;
          break;
        **end**
      **end**
    **end**
    **for** *j=1:$n_h$* **do**
      **while** *a finite number of trials* **do**
        $i = \lceil \mathcal{N}((j \times n_v)/n_h, \sigma^{neigh}) \rceil$;*%% sampled from a Gaussian distribution*;
        **if** $0 \leq i \leq n_v$ **then**
          addEdge $(i, j)$ to $G$;
          break;
        **end**
      **end**
    **end**
  **end**
  $L$ =computeAverageShorthestPath($G$);
**until** $L \leq L^{th}$;
*%% fit the topology to the data*;
re-arrange the visible nodes in $G$ s.t. the ones with higher degree correspond to data features with higher std. dev.;
*%% topology utilization*;
use the visible nodes from G as the visible layer in XBM (or GXBM);
use the hidden nodes from G as the hidden layer in XBM (or GXBM);
use the edges from G as the weights in XBM (or GXBM);

**Algorithm 1:** Pseudo-cod of the algorithm used to generate the topology of the XBM and GXBM models.

### 3.3 Complex Boltzmann machines

Just like restricted Boltzmann machines, complex Boltzmann machines are made up, by two layers of neurons, with connections only in between different layers and no connections within the same layer. The bottom layer (i.e. the visible one) is denoted further with a binary vector $\mathbf{v} = [v_1, v_2, .., v_{n_v}]$, in which each unit $v_i$ is a binary unit, and where $n_v$ is the size of $\mathbf{v}$ (i.e. the number of neurons of the visible layer). The top layer (i.e. the hidden one) is represented further by the binary vector $\mathbf{h} = [h_1, h_2, .., h_{n_h}]$, in which each element $h_j$ is binary, and where $n_h$ is the size of $\mathbf{h}$. Furthermore, each neuron from the visible layer has associated one bias. The biases of the visible neurons are collected in a vector $\mathbf{a} = [a_1, a_2, .., a_{n_v}]$. Similarly, hidden layer neurons have biases, collected in vector $\mathbf{b} = [b_1, b_2, .., b_{n_h}]$.

The difference between RBM and XBM consists in how the neurons from the different layers are connected between them. RBMs form a full undirected mesh between all the neurons on the hidden layer and all the neurons on the visible layer. By contrast, in XBMs the connections between the two layers are still undirected but sparse, as generated by Algorithm 1. Thus, XBMs have both scale-free and small-world topological properties. These connections are defined in a sparse adjacency weights matrix $\mathbf{W} = [[w_{11}, w_{12}, ..., w_{1n_h}], .., [w_{n_v1}, w_{n_v2}, ..., w_{n_vn_h}]]$ in which the elements are either null ($w_{ij} = 0$) when there is no connection between the visible neuron $i$ and the hidden neuron $j$ or have a connection weight ($w_{ij} \neq 0$) when the connection between $i$ and $j$ exists. The high level architectures of RBMs and XBMs are depicted in Figure 2. The sparse topology of XBMs leads to a much smaller number of connections, which further on leads to faster computational times than RBMs.

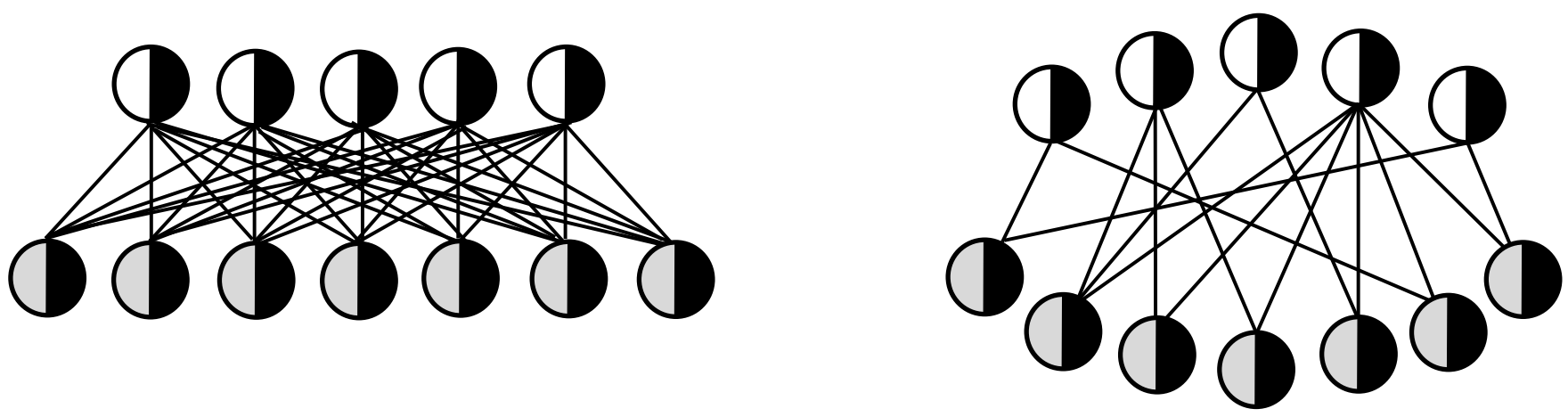

Fig. 2: Schematic architecture of: RBM (left) and XBM (right).

#### *3.3.1 The XBM energy function*

The energy function of an XBM is defined as:

$$E(v, h) = -\sum_{i=1}^{n_v} \sum_{j \in \mathbf{\Gamma}_i^h} v_i h_j w_{ij} - \sum_{i=1}^{n_v} v_i a_i - \sum_{j=1}^{n_h} h_j b_j \tag{4}$$

where, $\mathbf{\Gamma}_i^h$ is the set of all hidden neurons connected to the visible neuron $i$ (i.e. $\mathbf{\Gamma}_i^h = \{j | 1 \leq j \leq n_h, \forall j \in \mathbb{N} \wedge w_{ij} \neq 0\}$).

*3.3.2 XBM inference*

Due to the fact that there are no links between the neurons on the same layer, inference can still be performed in parallel for any visible neuron $i$ or any hidden neuron $j$, as below:

$$p(h_j = 1|\mathbf{v}, \mathbf{\Theta}) = \mathcal{S}\left(b_j + \sum_{i \in \mathbf{\Gamma}_j^v} v_i w_{ij}\right) \quad (5)$$

$$p(v_i = 1|\mathbf{h}, \mathbf{\Theta}) = \mathcal{S}\left(a_i + \sum_{j \in \mathbf{\Gamma}_i^h} h_j w_{ij}\right) \quad (6)$$

where, $\mathbf{\Gamma}_j^v$ is the set of all visible neurons connected to the hidden neuron $j$ (i.e. $\mathbf{\Gamma}_j^v = \{i|1 \leq i \leq n_v, \forall i \in \mathbb{N} \land w_{ij} \neq 0\}$), $\mathcal{S}(\cdot)$ is the sigmoid function, and $\mathbf{\Theta}$ represents the free parameters of the model (i.e. $\mathbf{W}$, $\mathbf{a}$, $\mathbf{b}$).

*3.3.3 XBM learning*

The general update rule for the free parameters $\mathbf{\Theta}$ of the GXBM model is given by:

$$\Delta\mathbf{\Theta}_{\tau+1} = \rho\Delta\mathbf{\Theta}_\tau + \alpha(\nabla\mathbf{\Theta}_{\tau+1} - \xi\mathbf{\Theta}_\tau) \quad (7)$$

where $\tau$, $\rho$, $\alpha$, and $\xi$ represent the update number, momentum, learning rate, and weights decay, respectively. For a thorough discussion on the optimal choice of these parameters the interested reader is referred to Hinton (2012). Furthermore, $\nabla\mathbf{\Theta}_{\tau+1}$ for each of the free parameters can be computed by using contrastive divergence (Hinton, 2002) and deriving the energy function from Equation 4 with respect to that parameter, yielding:

$$\nabla w_{ij} \propto \langle v_i h_j \rangle_0 - \langle v_i h_j \rangle_n; \forall i \in \mathbb{N}, 1 \leq i \leq n_v, \forall j \in \mathbf{\Gamma}_i^h; \quad (8)$$

$$\nabla a_i \propto \langle v_i \rangle_0 - \langle v_i \rangle_n; \forall i \in \mathbb{N}, 1 \leq i \leq n_v; \quad (9)$$

$$\nabla b_j \propto \langle h_j \rangle_0 - \langle h_j \rangle_n; \forall j \in \mathbb{N}, 1 \leq j \leq n_h; \quad (10)$$

with $\langle\cdot\rangle_n$ being the distribution of the model obtained after $n$ steps of Gibbs sampling in a Markov Chain which starts from the original data distribution $\langle\cdot\rangle_0$. We must note that in this paper we have chosen to train our proposed models using the original contrastive divergence method (Hinton, 2002), which is widely used and allows for a direct comparison to the results reported in the literature. It may well be that other training methods would offer better performance; yet, overall, we do not expect that a particular training method to significantly affect our findings.

## 3.4 Gaussian complex Boltzmann machines

Just like in GRBMs and RBMs, the only differences between GXBMs and XBMs is that in the case of GXBMs the visible layer $\mathbf{v} = [v_1, v_2, .., v_{n_v}]$ has real values and each $v_i$ is a linear unit with Gaussian noise (Hinton and Salakhutdinov, 2006). Thus, the total energy equation of GXBMs is slightly changed to reflect the real visible layer, as follows:

$$E(v,h) = -\sum_{i=1}^{n_v} \sum_{j \in \mathbf{\Gamma}_i^h} \frac{v_i}{\sigma_i} h_j w_{ij} - \sum_{i=1}^{n_v} \frac{(v_i - a_i)^2}{2\sigma_i^2} - \sum_{j=1}^{n_h} h_j b_j \quad (11)$$

where, $\sigma_i$ represents the standard deviation of the visible neuron $i$. We note that in the remainder we adopt the same notations used in XBM modeling, unless specified otherwise.

Furthermore, the inference in GXBMs can still be performed in parallel for any visible neuron $i$ or any hidden neuron $j$, as below:

$$p(h_j = 1|\mathbf{v}, \boldsymbol{\Theta}) = \mathcal{S}\left(b_j + \sum_{i \in \boldsymbol{\Gamma}_j^v} \frac{v_i}{\sigma_i} w_{ij}\right) \tag{12}$$

$$p(v_i = x|\mathbf{h}, \boldsymbol{\Theta}) = \mathcal{N}\left(a_i + \sum_{j \in \boldsymbol{\Gamma}_i^h} h_j w_{ij}, \sigma_i^2\right) \tag{13}$$

where, $\mathcal{N}(\cdot, \cdot)$ represents a Gaussian distribution. Finally, the learning in GXBM can be done using the same procedure as for XBM (Section 3.3.3).

## 4 Experimental results

To assess the performance of XBM and GXBM we have conducted three sets of experiments in a step-wise fashion. In the first one, we study the behavior of XBM, GXBM and their topology generation algorithm. In the second one, we analyze the reconstruction error obtained by GXBM on random generated data and on a real world dataset, more exactly on the Geographical Origin of Music dataset (Zhou et al, 2014). Thirdly, we assess the statistical performance of XBM on the MNIST digits dataset, CalTech 101 Silhouettes dataset (Marlin et al, 2010), and the 8 datasets from UCI evaluation suite (Larochelle and Murray, 2011) using Annealed Importance Sampling (AIS) (Salakhutdinov and Murray, 2008).

Furthermore, in the last two sets of experiments, we compare GXBM/XBM against three other methods, as follows: (1) the standard fully connected GRBM/RBM; (2) sparse GRBM/RBM models, denoted further $\text{GRBM}_{\text{FixProb (Fixed Probability)}}/\text{RBM}_{\text{FixProb}}$, in which the probability for any possible connection to exist is set to the number of weights of the counterpart GXBM/XBM model divided by the total number of possible connection for that specific configuration of hidden and visible neurons[4]; and (3) sparse GRBM/RBM models, denoted further $\text{GRBM}_{\text{TrPrTr (Train Prune Train)}}/\text{RBM}_{\text{TrPrTr}}$, in which the sparsity is obtained using the algorithm introduced in Han et al (2015) with L2 regularization, and in which the weights sparsity target is set to the number of weights of the counterpart GXBM/XBM model. Please note that in all experiments if the weights sparsity target was not reached after 50 pruning iterations, we stopped the training algorithm and we used the obtained $\text{GRBM}_{\text{TrPrTr}}/\text{RBM}_{\text{TrPrTr}}$ model. For each evaluated case, we analyze two scenarios when: (1) the number of connections is the same for the sparse and the full connected models, while the number of hidden neurons is different; (2) the number of hidden neurons is the same for the sparse and the full connected models, while the number of connections is different.

### 4.1 Scrutinizing XBM and GXBM topologies

In this set of experiments, we analyze the characteristics of the XBM sparse topology. For the sake of brevity, we refer just to the XBM - RBM relation, since the GXBM - GRBM is identical from the topological point of view. To perform the various operations on the bipartite graph we have used the NetworkX library (Hagberg et al, 2008), setting the power law coefficient $\gamma$ to 2, a typical value in various real-world networks (Clauset et al, 2009). Firstly, we verified the output of the topology-generation algorithm (i.e. Algorithm 1). Being one of the algorithm constraints, the small-world property is preserved in XBM. At the same time, due to the fact that the neurons from different layers are connected starting from a power law degree sequence, the scale-free property is also preserved. As example, Figure 3a depicts the weights distribution for an XBM having 784 visible and 1000 hidden neurons, while Figure 3b shows the degree distribution of an XBM with 100 visible and 1000 hidden neurons on the loglog scale. Figure 3b exhibits evidently a scale-free degree distribution. All other experiments exhibited the required scale-free distribution. Furthermore, we analyzed the number of connections in XBM in comparison with the number of connections in RBMs, given the same number of hidden ($n_h$) and visible ($n_v$) neurons. Figure 3c depicts how many times the number of connections in RBM is bigger than the number of connections in XBM for various configurations (the number of hidden and visible neurons varies from 10 to 1000). The actual values in the heat map are computed using the following formula $n_w^{RBM}/n_w^{XBM}$, where $n_w^{XBM}$ is obtained after counting the links given by the topology generation algorithm for XBM, and $n_w^{RBM} = n_v n_h$. It can be observed that as the number of hidden and visible neurons increases, the number of weights in XBM becomes smaller and smaller than the one in RBM. For instance, we achieve around 14 times less weights in XBM for 100 visible and 100 hidden neurons, and approximatively 95 times less weights in XBM for 1000 visible and 1000 hidden neurons.

[4] Please note that this procedure yields approximately the same number of connections in $\text{GRBM}_{\text{FixProb}}/\text{RBM}_{\text{FixProb}}$ as in GXBM/XBM to ensure a fair comparison.

### 4.2 GXBM evaluation

In the second set of experiments, we assess the performance of GXBM against GRBM, $\text{GRBM}_{\text{FixProb}}$, and $\text{GRBM}_{\text{TrPrTr}}$ on randomly generated as well as on the real-world dataset Geographical Origin of Music (Zhou et al, 2014).

*Settings and implementations.* For all experiments performed in this set we have used Python implementations of the four models under scrutiny. In all models, the momentum was set to 0.5, the learning rate to 0.001, and the number of Gibbs sampling in contrastive divergence to 1, as discussed by Hinton (2012). The weights decay was 0.0002 for GXBM, GRBM, and $\text{GRBM}_{\text{FixProb}}$, while for $\text{GRBM}_{\text{TrPrTr}}$ we used the L2 regularization. The number of neurons in the visible layer was set to the dimension of the input data, while the number of hidden neurons was varied for a better comparison. In the learning phase, we stopped the models after 100 training epochs to ensure full convergence. In fact, convergence was much faster, as exemplified in Figure 3d which shows the case of GXBM with 100 visible and 1000 hidden neurons trained on random generated data. In the case of $\text{GRBM}_{\text{TrPrTr}}$, we have repeated the training procedure for a maximum of 50 pruning iterations trying to reach the same amount of weights as in GXBM, but this target was impossible to reach in all situations.

*Performance metrics.* To quantify the performance of the models, we used a variety of standard metrics. We have used: the Root Mean Square Error (RMSE) to estimate the distance between the reconstructed inputs and the ground truth; the Pearson Correlation Coefficient (PCC) to reflect the correlations between the estimated inputs and the ground truth; and the P-value to arrive at a statistically significant reconstruction level during the learning phase.

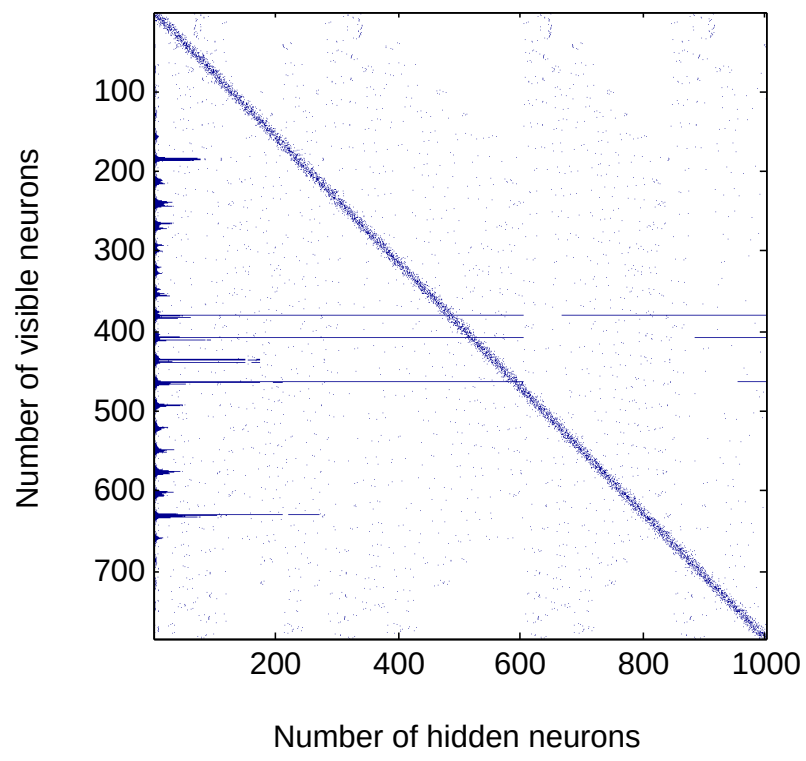

(a) Fig. 3a: Example of weight distribution for an XBM with 784 visible and 1000 hidden neurons.

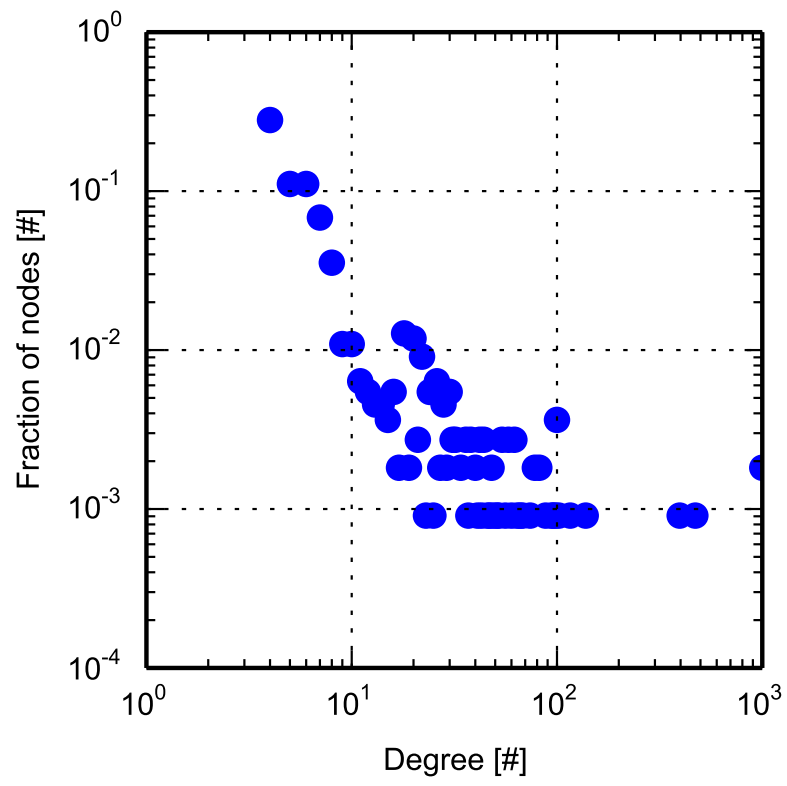

(b) Fig. 3b: The degree distribution of an XBM with 1000 hidden neurons and 100 visible ones (loglog scale).

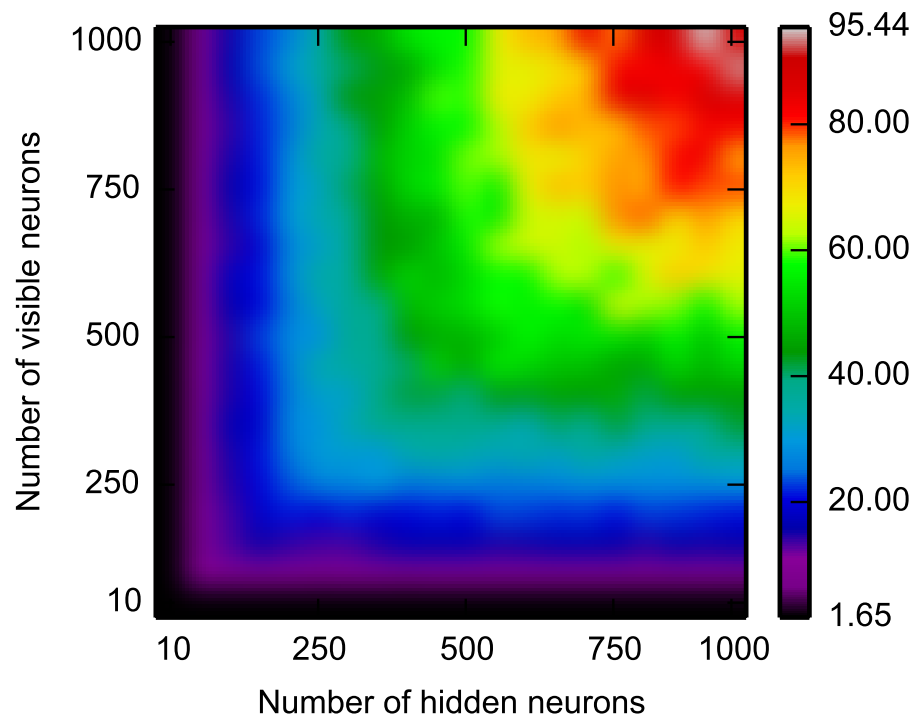

(c) Fig. 3c: Studying the relation between the number of weights in RBM and XBM (the heatmap values are given by $n_w^{RBM}/n_w^{XBM}$).

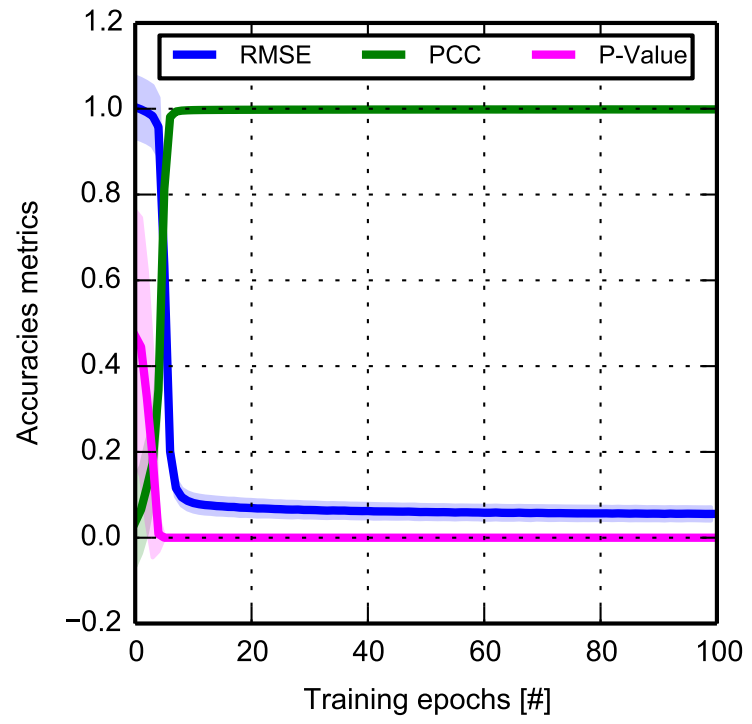

(d) Fig. 3d: Training behavior on random data for an GXBM with 1000 hidden neurons and 100 visible ones.

#### *4.2.1 GXBM performance on random generated data*

Firstly, we analyze how well GXBM is capable to reconstruct random generated data. To this end, we have generated 1000 data points, each one having 100 dimensions, and each dimension being sampled from a Gaussian distribution with 0 mean and standard deviation equal to $1$, i.e. $\mathcal{N}(0,1)$. Due to the fact that these are random generated data, there was no reason to use cross validation. Thus, we have used $70\%$ of data to train the models, and the remaining $30\%$ to test the models. Firstly, we analyzed the reconstruction capabilities of GXBM, $\text{GRBM}_{\text{FixProb}}$, $\text{GRBM}_{\text{TrPrTr}}$, and GRBM, given the the same number of weights. Figure 4 (left) depicts this situation, while the number of weights were varied from 700 up to approximately $7000$. Clearly, GXBM outperforms GRBM in both, RMSE and PCC, while its internal topology permits it to have a higher number of hidden neurons. Remarkably, with approximately $1000$ weights, the mean RMSE for GXBM is already very low, around $0.3$, while the mean PCC is almost perfect, over $0.95$. By contrast, GRBM with $1000$ weights performed poorly. Furthermore, it is clear that the GRBM performance increases with the number of weights, yet it can be observed that even at approximately $7000$ weights GRBM is not capable to reach the same level of performance of the GXBM with 1000 weights. Besides that, GXBM outperforms also the other sparse models, GRBM$_{\text{FixProb}}$ and GRBM$_{\text{TrPrTr}}$, but not so drastically. In fact, GRBM$_{\text{FixProb}}$ has very close performance to GXBM. GRBM$_{\text{TrPrTr}}$ is not so close, while having also a very high computational cost as depicted in the fourth row of Figure 4.

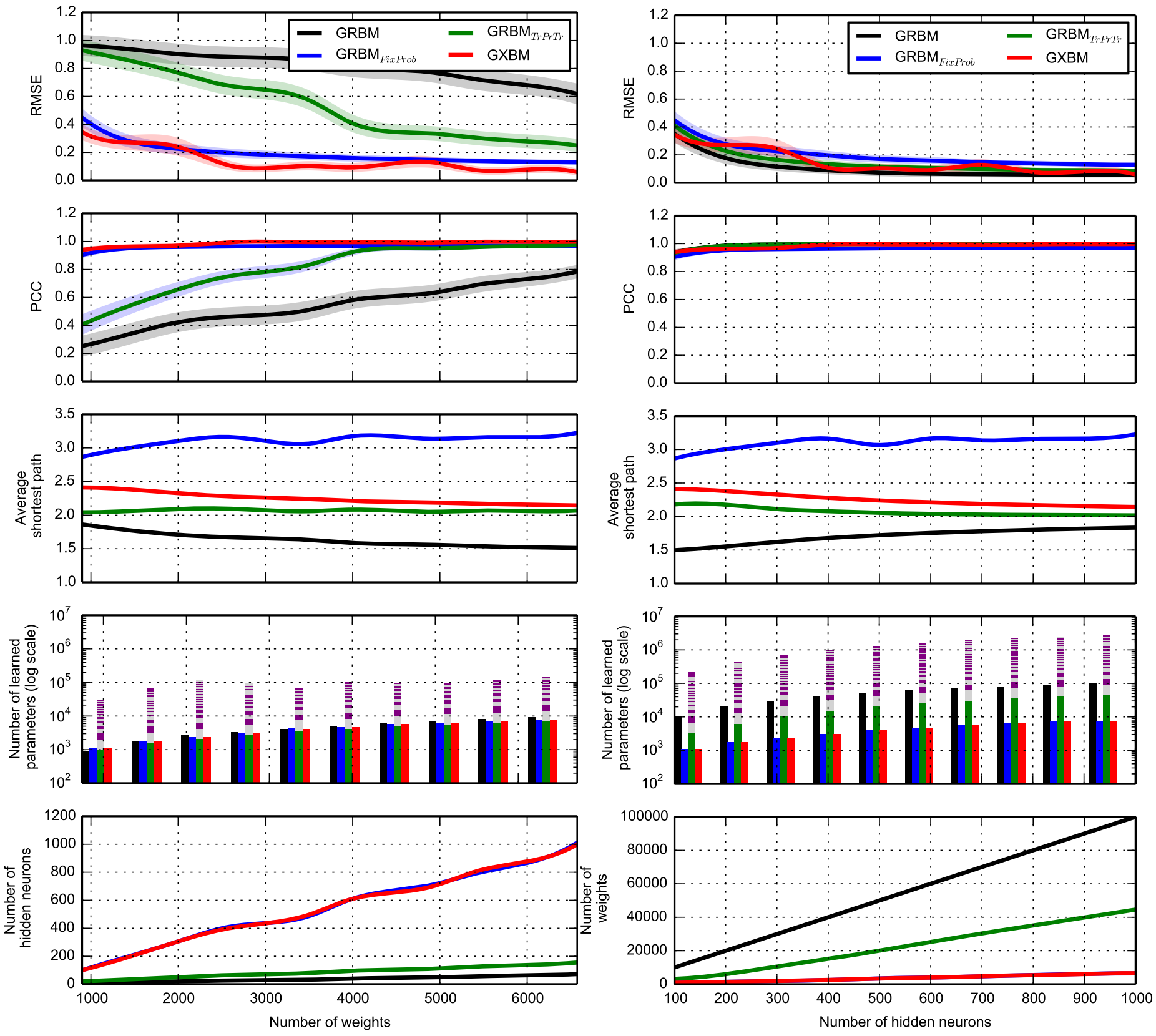

Fig. 4: Reconstruction capability on random generated data of GRBM, GRBM$_{\text{FixProb}}$, GRBM$_{\text{TrPrTr}}$ and GXBM with: (left) the same number of weights; (right) the same number of hidden neurons. The straight line represents the mean; the shadowed area shows the standard deviation. The number of learned parameters for GRBM$_{\text{TrPrTr}}$ is given in green after the last pruning iteration, while above the green color the alternating gray and purple colors represent the number of learned parameters at each pruning iteration starting with the first one from the top.

To better understand these differences, we proceed to the next scenario, analyzing GRBM, GRBM$_{\text{FixProb}}$, GRBM$_{\text{TrPrTr}}$, and GXBM having the same number of hidden neurons. This is reflected in Figure 4 (right) in which the number of hidden neurons is varied from 100 to 1000. Surprising, even though the number of free parameters (i.e. weights) was smaller by at least one order of magnitude in GXBM (as it can be seen in the bottom-right plot of Figure 4), GXBM performs similarly to GRBM in terms of PCC and RMSE, while GRBM$_{\text{FixProb}}$ and GRBM$_{\text{TrPrTr}}$ reach almost a similar performance. Still, GRBM$_{\text{TrPrTr}}$ has the downside of not being capable to reach the same number of weights as GXBM or GRBM$_{\text{FiXProb}}$ even after 50 prunning iterations, as reflected on the fourth and fifth rows of Figure 4. Interestingly, for this specific dataset, all models seem to reach their maximum learning capacity when they have approximately 400 hidden neurons, showing no further improvement after this point.

#### *4.2.2 GXBM performance on geographical ethnomusicology data*

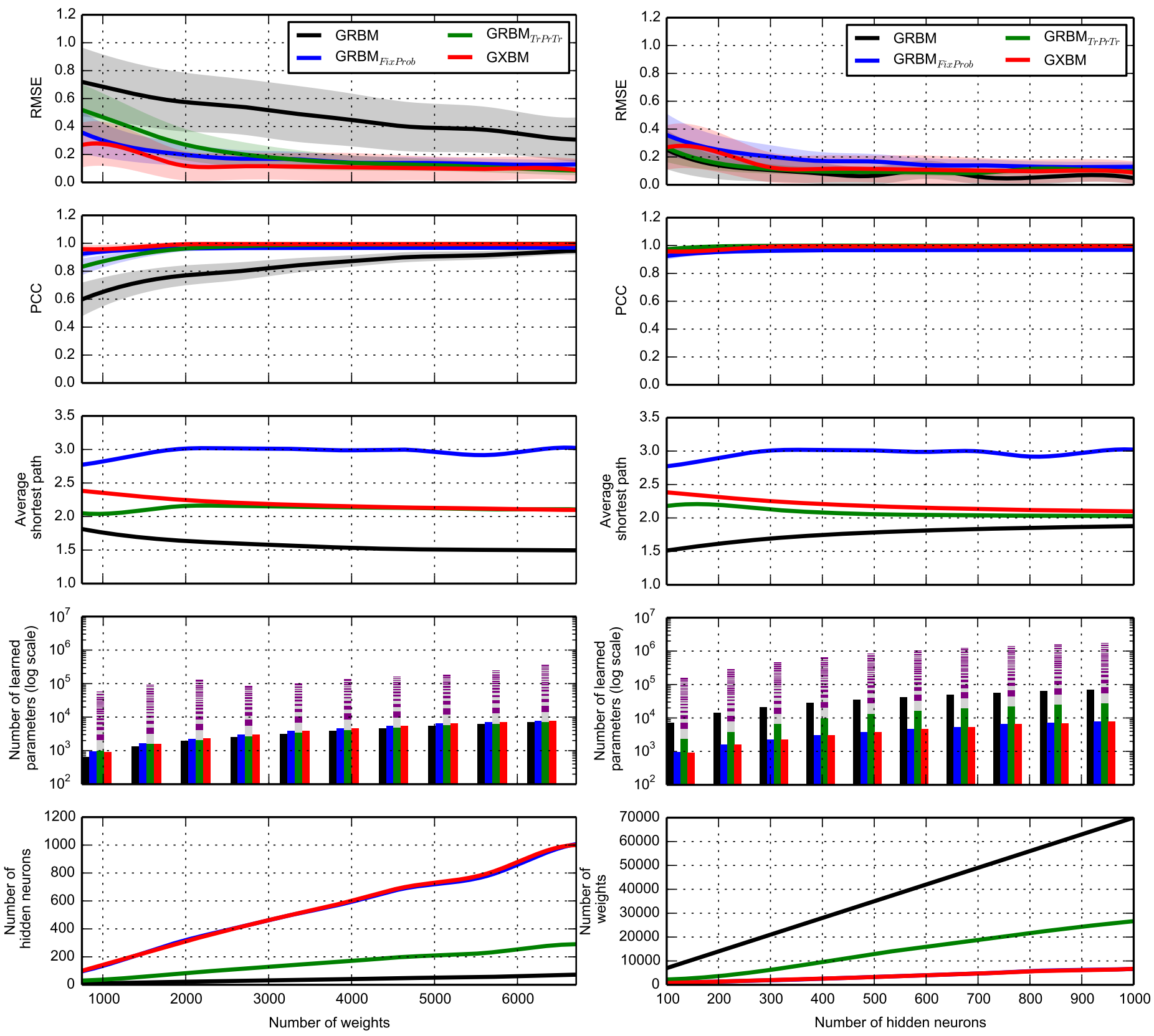

Fig. 5: Reconstruction capabilities on the geographical ethnomusicology dataset of GRBM, GRBM$_{\mathrm{FixProb}}$, GRBM$_{\mathrm{TrPrTr}}$, and GXBM with: (left) the same number of weights; (right) the same number of hidden neurons. The straight line represents the mean; the shadowed area shows the standard deviation. The number of learned parameters for GRBM$_{\mathrm{TrPrTr}}$ is given in green after the last pruning iteration, while above the green color the alternating gray and purple colors represent the number of learned parameters at each pruning iteration starting with the first one from the top.

We have then assessed the reconstruction capabilities of GXBM on a real world dataset. We have used the Geographical Origin of Music dataset (Zhou et al, 2014). This contains 1059 tracks from different countries, each track having 70 dimensions (i.e. the first 68 represent audio features, while the last ones are latitude and longitude of each specific song), already normalized to have mean 0 and standard deviation equal to 1. We performed a 10-fold cross validation. The averaged results depicted in Figure 5 confirm the same findings obtained on the random generated dataset (Section 4.2.1). Given the same number of weights, GXBM outperforms clearly GRBM, and outperforms slightly $\text{GRBM}_{\text{FixProb}}$ and $\text{GRBM}_{\text{TrPrTr}}$, while the four perform similarly when the number of hidden neurons is comparable. By analyzing the reconstruction performance metrics (i.e. RMSE and PCC) reported in Figure 5 and Figure 4 it is interesting to observe that GXBM and $\text{GRBM}_{\text{FixProb}}$ are more stable, independently if the data are random or non-random. By contrast, GRBM and $\text{GRBM}_{\text{TrPrTr}}$ performance depend more on the data type.

To assess GXBM from a different perspective, we performed a small regression experiment on the geographical ethnomusicology dataset, even though the regression task does not constitute one of the goals of this paper. Thus, we compared the ability of GRBM and GXBM in predicting the latitude and the longitude corresponding to the 68 audio features across all the tracks. We can see from Figure 5 that the dataset is well represented with just about 400 hidden neurons; thus we have used this value in both models. In addition, in GXBM we have set the visible neurons corresponding to latitude and longitude to the first two visible neurons having the largest number of connections. We then performed a 10-fold cross validation to obtain the average distance error between the predicted latitude and longitude and their true value counterparts. The resulting predictions were $3258 \pm 175$ kilometers (GRBM) and $3252 \pm 176$ kilometers (GXBM) which are comparable to the top performers found in Zhou et al (2014), even if it is well known that GRBMs and RBMs are not best performers on classification and regression tasks, when used as standalone models - best performance would pursued by stacking these models in deep architectures (Bengio, 2009).

From the third row of Figures 4 and 5, it is very interesting to observe that, even if it was not a target, the random generated connections of $\text{GRBM}_{\text{FixProb}}$ exhibit a very small average shortest path. This observation may explain the good performance of $\text{GRBM}_{\text{FixProb}}$ close to the one of GXBM in this set of experiments, while having a similar computational time (i.e. the same number of weights to be computed). Even more, the $\text{GRBM}_{\text{TrPrTr}}$ models end up still having a small-world topology after the iterative pruning process. Still, the better performance obtained by GXBM is given by, all-together, its scale-free topology, the local neighborhoods connections, and the consideration of data distribution in the topology.

### 4.3 XBM evaluation

In the third set of experiments, we have assessed the performance of XBM on the MNIST digits dataset[5], CalTech 101 Silhouettes dataset (Marlin et al, 2010), and the 8 datasets from the UCI evaluation suite (Larochelle and Murray, 2011).

*Settings, implementations, and performance metrics.* To allow for a direct comparison, we have adopted the same settings for all datasets. This time, all the models were implemented and the experiments were performed using the MATLAB® environment, partially to facilitate the comparisons with the results reported on the MNIST dataset in Salakhutdinov and Murray (2008). For each RBM model, we have trained two XBM, two $\text{RBM}_{\text{FixProb}}$, and two $\text{RBM}_{\text{TrPrTr}}$ models, as follows: (1) one having the same number of hidden neurons, but with much fewer weights; (2) the other with approximatively the same number of weights but with a higher number of hidden neurons. Just as in Salakhutdinov and Murray (2008), we have used a fixed learning rate (i.e. 0.05) for all datasets, with the exception of the MNIST dataset when a decreasing learning rate was used for the situation in which the number of contrastive divergence steps was gradually increased from 1 up to 25, as suggested in Carreira-Perpinan and Hinton (2005). In all cases, the number of training epochs was set to 259 and the training and testing data were split in mini-batches of 100 samples. Finally, to assess the performance of XBM, $RBM_{FixProb}$, and $RBM_{TrPrTr}$, we have used AIS (adopting the very same settings as in Salakhutdinov and Murray (2008)) to estimate the average log-probabilities on the training and testing data of the datasets.

[5] http://yann.lecun.com/exdb/mnist/, Last visit on October $18^{th}$ 2015.

### *4.3.1 XBM performance on the MNIST digits dataset*

The MNIST digits dataset is widely used to assess the performance of novel machine learning algorithms. The dataset contains 60000 images for training and 10000 images for testing, each image being a variant of a digit represented by a 28x28 binary matrix.

The results depicted in Table 1[6] confirm the findings from the previous set of experiments (i.e. GXBM - GRBM case). We see that: (1) given the same number of hidden neurons an XBM model has a generative performance close to an RBM, yet having the significant benefit of much fewer weights (i.e. this offers much smaller computational time); (2) given approximatively the same number of weights (i.e. at comparable computational time) an XBM model has a better generative performance than an RBM model. The results suggest that XBM models can achieve good generative performance already with one step contrastive divergence and a fix learning rate. This is because the performance gain obtained by increasing the number of contrastive divergence steps is smaller than in the case of RBMs. This may be considered also an advantage for XBMs as it is a common practice to use just one step contrastive divergence to decrease the learning time. It is worth highlighting, that an XBM with 15702 weights reaches 36.02 *nats* better than an RBM with 15680 weights and 20 hidden neurons. Similarly to the GXBM behaviour, as the models increase in size, we observe that the difference between XBMs and RBMs gets smaller. For instance, an XBM with 387955 weights is 10.89 *nats* better than an RBM with 392000 weights, which, further on, is just 6.99 *nats* better than an XBM having the same number of hidden neurons (i.e. 500) but with approximately 40 times fewer weights (i.e. 10790). Also worth noting that when the CD learning steps were gradually increased from 1 to 25 the XBM model with 387955 weights slightly outperformed (i.e. 1.13 *nats*) the RBM model having 500 hidden neurons and 392000 weights trained in Salakhutdinov and Murray (2008). We should note that the latter is considered to be one of the best generative RBM models reported in the literature, to the best of our knowledge.

Finally, we would like to highlight that in 5 out of the 6 cases considered, XBMs outperform all the other models including the sparse ones, while in the remaining case, the best performer is not the fully connected RBM as expected, but still a sparse model,i.e. $RBM_{TrPrTr}$. Yet, the latter one, same as in Subsection 4.2, shows some robustness issues as it performs very badly for a large number of hidden neurons (i.e. 27000), while its computational time is much higher than for all the other models considered. Remarkably, $RBM_{FixProb}$ obtains very good results in all cases, being behind XBM just from few up to a maximum of approximately 25 *nats* in the worst case.

[6] The average cluster coefficient was computed using the method proposed for bipartite graphs in Latapy et al (2008).

Table 1: Estimation of the average log-probabilities on the training and testing data obtained from the MNIST digits dataset using AIS. The results for RBMs are taken from Salakhutdinov and Murray (2008).

| No. of CD steps during learning | No. of weights | Model | No. of hidden units | Average shortest path | Average cluster coefficient | No. of pruning iterations | Average train log-probabilities | Average test log-probabilities |
|---|---|---|---|---|---|---|---|---|
| 1 (fixed) | 15680 | RBM | 20 | 1.94 | 1 | 0 | -164.87 | -164.50 |
| | 15702 | XBM | 1000 | 2.70 | 0.089 | 0 | -130.26 | **-128.48** |
| | 15730 | $RBM_{FixProb}$ | 1000 | 3.19 | 0.034 | 0 | -143.32 | -142.34 |
| | 65583 | $RBM_{TrPrTr}$ | 1000 | 2.45 | 0.060 | 50 | -146.51 | -147.47 |
| | 4954 | XBM | 20 | 2.01 | 0.396 | 0 | -167.36 | -166.60 |
| | 4896 | $RBM_{FixProb}$ | 20 | 2.25 | 0.209 | 0 | -169.55 | -169.10 |
| | 6119 | $RBM_{TrPrTr}$ | 20 | 2.22 | 0.276 | 50 | -176.50 | -176.20 |
| | 19600 | RBM | 25 | 1.93 | 1 | | -153.46 | -152.68 |
| | 19527 | XBM | 1500 | 2.70 | 0.071 | 0 | -126.74 | **-126.07** |
| | 19875 | $RBM_{FixProb}$ | 1500 | 3.28 | 0.037 | 0 | -143.83 | -142.91 |
| | 103395 | $RBM_{TrPrTr}$ | 1500 | 2.38 | 0.059 | 50 | -148.49 | -151.08 |
| | 6358 | XBM | 25 | 2.01 | 0.350 | 0 | -163.39 | -161.09 |
| | 6389 | $RBM_{FixProb}$ | 25 | 2.12 | 0.205 | 0 | -162.66 | -162.02 |
| | 6593 | $RBM_{TrPrTr}$ | 25 | 2.25 | 0.234 | 23 | -170.29 | -169.68 |
| | 392000 | RBM | 500 | 1.52 | 1 | 0 | -122.86 | -125.53 |
| | 387955 | XBM | 27000 | 2.05 | 0.156 | 0 | -115.28 | **-114.64** |
| | 391170 | $RBM_{FixProb}$ | 27000 | 2.87 | 0.053 | 0 | -140.97 | -140.30 |
| | 2204393 | $RBM_{TrPrTr}$ | 27000 | 2.10 | 0.071 | 50 | -602.94 | -652.21 |
| | 10790 | XBM | 500 | 2.44 | 0.082 | 0 | -134.06 | -132.52 |
| | 10846 | $RBM_{FixProb}$ | 500 | 3.12 | 0.039 | 0 | -149.27 | -148.42 |
| | 29616 | $RBM_{TrPrTr}$ | 500 | 2.62 | 0.064 | 50 | -131.48 | -131.34 |
| 3 (fixed) | 19600 | RBM | 25 | 1.93 | 1 | 0 | -144.11 | -143.20 |
| | 19527 | XBM | 1500 | 2.70 | 0.071 | 0 | -122.19 | -121.59 |
| | 19875 | $RBM_{FixProb}$ | 1500 | 3.28 | 0.037 | 0 | -139.89 | -138.86 |
| | 142237 | $RBM_{TrPrTr}$ | 1500 | 2.34 | 0.074 | 50 | -116.11 | **-120.50** |
| | 6358 | XBM | 25 | 2.01 | 0.350 | 0 | -158.67 | -157.69 |
| | 6389 | $RBM_{FixProb}$ | 25 | 2.12 | 0.205 | 0 | -161.35 | -160.69 |
| | 7715 | $RBM_{TrPrTr}$ | 25 | 2.14 | 0.259 | 50 | -159.05 | -158.39 |
| | 392000 | RBM | 500 | 1.52 | 1 | 0 | -102.81 | -105.50 |
| | 387955 | XBM | 27000 | 2.05 | 0.156 | 0 | -103.66 | **-101.93** |
| | 391170 | $RBM_{FixProb}$ | 27000 | 2.87 | 0.053 | 0 | -125.51 | -125.03 |
| | 2827787 | $RBM_{TrPrTr}$ | 27000 | 2.05 | 0.087 | 5 | -488.07 | -512.56 |
| | 10790 | XBM | 500 | 2.44 | 0.082 | 0 | -128.07 | -127.41 |
| | 10846 | $RBM_{FixProb}$ | 500 | 3.12 | 0.039 | 0 | -145.11 | -144.08 |
| | 38991 | $RBM_{TrPrTr}$ | 500 | 2.46 | 0.069 | 50 | -112.27 | -112.45 |
| from 1 to 25 (variable) | 392000 | RBM | 500 | 1.52 | 1 | 0 | -83.10 | -86.34 |
| | 387955 | XBM | 27000 | 2.05 | 0.156 | 0 | -86.12 | **-85.21** |
| | 391170 | $RBM_{FixProb}$ | 27000 | 2.87 | 0.053 | 0 | -107.23 | -106.78 |
| | 3262957 | $RBM_{TrPrTr}$ | 27000 | 2.18 | 0.076 | 50 | -349.87 | -376.92 |
| | 10790 | XBM | 500 | 2.44 | 0.082 | 0 | -121.26 | -120.43 |
| | 10846 | $RBM_{FixProb}$ | 500 | 3.12 | 0.039 | 0 | -136.27 | -135.89 |
| | 36674 | $RBM_{TrPrTr}$ | 500 | 2.35 | 0.071 | 50 | -134.25 | -135.76 |

#### *4.3.2 XBM performance on the CalTech 101 Silhouettes dataset*

To confirm the previous results on a different (more complicated) dataset, further on we assess XBMs generative performance on the CalTech 101 Silhouettes dataset (Marlin et al, 2010). This dataset contains silhouettes of objects extracted from the CalTech 101 image dataset. In total it has 101 classes and two datasets. One with binary images of 28x28 pixels split in a training set of 4100 samples and a testing set of 2307 samples, and one with binary images of 16x16 pixels split in a training set of 4082 samples and a testing set of 2302 samples. As our goal was not to fine tune the four models, but to have a clear direct comparison between them, for each dataset we have used 1 CD step and we considered two evaluation cases, a small RBM (i.e. 25 hidden neurons) and a large one (i.e. 500 hidden neurons). Table 2 confirms our previous findings and shows that XBMs are still the best performers outperforming clearly all the other models, with a striking difference of 118.48 *nats* against fully connected RBMs (which are subject to over-fitting) on the dataset of 28x28 image size. On both datasets, it is interesting to see that the best XBMs performers are not the largest models (i.e. 27000 and 10000 hidden neurons, respectively), but the average size ones (i.e. 1500 and 500 hidden neurons, respectively).

Table 2: Estimation of the average log-probabilities on the training and testing data obtained from the CalTech 101 Silhouettes dataset using 1 step CD and AIS.

| **Dataset** | **No. of weights** | **Model** | **No. of hidden units** | **Average shortest path** | **Average cluster coefficient** | **No. of pruning iterations** | **Average train log-probabilities** | **Average test log-probabilities** |
|---|---|---|---|---|---|---|---|---|
| | 19600 | RBM | 25 | 1.93 | 1 | 0 | -329.74 | -315.22 |
| | 19201 | XBM | 1500 | 2.91 | 0.099 | 0 | -146.97 | **-142.96** |
| | 19979 | $RBM_{FixProb}$ | 1500 | 3.28 | 0.037 | 0 | -164.84 | -162.41 |
| | 414478 | $RBM_{TrPrTr}$ | 1500 | 2.13 | 0.223 | 50 | -141.77 | -360.24 |
| | 6423 | XBM | 25 | 2.02 | 0.47 | 0 | -330.37 | -323.01 |
| | 6341 | $RBM_{FixProb}$ | 25 | 2.14 | 0.20 | 0 | -356.68 | -353.25 |
| 28x28 | 6531 | $RBM_{TrPrTr}$ | 25 | 2.04 | 0.316 | 13 | -350.66 | -339.94 |
| image size | 392000 | RBM | 500 | 1.52 | 1 | 0 | -161.05 | -261.44 |
| | 392464 | XBM | 27000 | 2.05 | 0.136 | 0 | -178.16 | -187.49 |
| | 381228 | $RBM_{FixProb}$ | 27000 | 3.51 | 0.040 | 0 | -277.06 | -283.15 |
| | 390410 | $RBM_{TrPrTr}$ | 27000 | 3.02 | 0.017 | 13 | -145.38 | -307.14 |
| | 11077 | XBM | 500 | 2.46 | 0.094 | 0 | -169.56 | **-164.07** |
| | 11350 | $RBM_{FixProb}$ | 500 | 3.09 | 0.038 | 0 | -196.24 | -191.71 |
| | 34479 | $RBM_{TrPrTr}$ | 500 | 2.15 | 0.212 | 50 | -240.09 | -403.63 |
| | 6400 | RBM | 25 | 1.83 | 1 | 0 | -102.87 | -94.88 |
| | 6359 | XBM | 500 | 2.40 | 0.095 | 0 | -74.60 | **-69.95** |
| | 6364 | $RBM_{FixProb}$ | 500 | 2.93 | 0.048 | 0 | -94.68 | -89.93 |
| | 44573 | $RBM_{TrPrTr}$ | 500 | 2.13 | 0.211 | 50 | -121.19 | -166.37 |
| | 2296 | XBM | 25 | 2.04 | 0.400 | 0 | -99.78 | -93.64 |
| | 2321 | $RBM_{FixProb}$ | 25 | 2.09 | 0.226 | 0 | -100.18 | -92.89 |
| 16x16 | 2084 | $RBM_{TrPrTr}$ | 25 | 2.03 | 0.350 | 3 | -114.54 | -106.69 |
| image size | 128000 | RBM | 500 | 1.54 | 1 | 0 | -70.89 | -98.64 |
| | 123580 | XBM | 10000 | 2.04 | 0.191 | 0 | -77.96 | -78.43 |
| | 122841 | $RBM_{FixProb}$ | 10000 | 3.09 | 0.055 | 0 | -104.06 | -102.90 |
| | 609307 | $RBM_{TrPrTr}$ | 10000 | 2.09 | 0.114 | 50 | -102.48 | -101.16 |
| | 6721 | XBM | 500 | 2.70 | 0.147 | 0 | -73.83 | **-69.29** |
| | 6407 | $RBM_{FixProb}$ | 500 | 2.92 | 0.048 | 0 | -83.37 | -78.48 |
| | 44573 | $RBM_{TrPrTr}$ | 500 | 2.13 | 0.211 | 50 | -121.19 | -166.37 |

Table 3: The characteristics of the UCI evaluation suite datasets.

| **Dataset** | **No. of inputs** | **Training set size** | **Testing set size** |
|---|---|---|---|
| Adult | 123 | 5000 | 26147 |
| Connect4 | 126 | 16000 | 47557 |
| DNA | 180 | 1400 | 1186 |
| Mushrooms | 112 | 2000 | 5624 |
| NIPS-0-12 | 500 | 400 | 1240 |
| OCR-letters | 128 | 32152 | 10000 |
| RCV1 | 150 | 40000 | 150000 |
| Web | 300 | 14000 | 32561 |

#### *4.3.3 XBM performance on the UCI evaluation suite*

Up to now all the datasets used to evaluate the XBMs were binarized images. In this last subset of experiments, we use the UCI evaluation suite, which contains 8 binary datasets coming from various domains. These datasets are carefully selected by Larochelle and Murray (2011) to assess the performance of generative and density estimation models. Their characteristics are well described in Germain et al (2015) and are summarized in Table 3. As before, to have a clear direct comparisons of all sparse models we have compared them with baseline fully connected RBMs with 100 hidden neurons using standard 1 CD step. Table 4 summarizes the results. XBMs outperform all the other models, including the fully connected RBMs, on 7 out of the 8 datasets. As usual, $RBM_{FixProb}$ shows a good performance overall, being even the best performer on one dataset, i.e. Connect4. By contrast, $RBM_{TrPrTr}$ has robustness issues, sometimes showing good generative capabilities and sometimes not. Even if it was not in our goal to outperform the best results from the literature and, as a consequence, we did not fine tune any parameter and we did not try other training algorithms, XBMs reach on all datasets very good performances close to the ones of the best generative models carefully optimized in Germain et al (2015).

Table 4: Estimation of the average log-probabilities on the training and testing data obtained from the UCI evaluation suite datasets using using 1 step CD and AIS.

| Dataset | No. of weights | Model | No. of hidden units | Average shortest path | Average cluster coefficient | No. of pruning iterations | Average train log-probabilities | Average test log-probabilities |
|---|---|---|---|---|---|---|---|---|
| | 12300 | RBM | 100 | 1.49 | 1 | 0 | -17.56 | -17.86 |
| | 12911 | XBM | 1200 | 2.35 | 0.154 | 0 | -15.51 | **-15.89** |
| | 12692 | $RBM_{FixProb}$ | 1200 | 2.80 | 0.072 | 0 | -17.31 | -17.56 |
| Adult | 11211 | $RBM_{TrPrTr}$ | 1200 | 2.56 | 0.071 | 4 | -135.29 | -135.98 |
| | 1617 | XBM | 100 | 2.36 | 0.129 | 0 | -17.92 | -17.97 |
| | 1641 | $RBM_{FixProb}$ | 100 | 2.50 | 0.082 | 0 | -18.95 | -19.04 |
| | 2089 | $RBM_{TrPrTr}$ | 100 | 2.28 | 0.348 | 2 | -100.20 | -100.40 |
| | 12600 | RBM | 100 | 1.49 | 1 | 0 | -16.80 | -17.00 |
| | 12481 | XBM | 1200 | 2.14 | 0.142 | 0 | -17.27 | -17.37 |
| | 12498 | $RBM_{FixProb}$ | 1200 | 2.83 | 0.072 | 0 | -15.13 | **-15.23** |
| Connect4 | 51412 | $RBM_{TrPrTr}$ | 1200 | 2.10 | 0.211 | 50 | -17.63 | -18.11 |
| | 1692 | XBM | 100 | 2.36 | 0.164 | 0 | -25.63 | -25.68 |
| | 1722 | $RBM_{FixProb}$ | 100 | 2.48 | 0.083 | 0 | -32.01 | -32.03 |
| | 1922 | $RBM_{TrPrTr}$ | 100 | 2.43 | 0.188 | 6 | -41.43 | -41.52 |
| | 18000 | RBM | 100 | 1.53 | 1 | 0 | -94.75 | -99.52 |
| | 17801 | XBM | 1600 | 2.71 | 0.157 | 0 | -79.05 | **-83.17** |
| | 18314 | $RBM_{FixProb}$ | 1600 | 2.93 | 0.060 | 0 | -78.57 | -85.53 |
| DNA | 17597 | $RBM_{TrPrTr}$ | 1600 | 3.26 | 0.087 | 13 | -143.77 | -155.75 |
| | 2267 | XBM | 100 | 2.41 | 0.133 | 0 | -89.31 | -90.31 |
| | 2291 | $RBM_{FixProb}$ | 100 | 2.51 | 0.079 | 0 | -91.53 | -92.98 |
| | 2231 | $RBM_{TrPrTr}$ | 100 | 2.44 | 0.177 | 4 | -111.45 | -114.17 |
| | 11200 | RBM | 100 | 1.49 | 1 | 0 | -24.77 | -25.60 |
| | 10830 | XBM | 1000 | 2.14 | 0.156 | 0 | -14.21 | **-14.71** |
| | 10639 | $RBM_{FixProb}$ | 1000 | 2.73 | 0.075 | 0 | -15.29 | -15.82 |
| Mushrooms | 22376 | $RBM_{TrPrTr}$ | 1000 | 2.26 | 2.26 | 50 | -21.76 | -23.09 |
| | 1515 | XBM | 100 | 2.39 | 0.11 | 0 | -17.14 | -17.54 |
| | 1451 | $RBM_{FixProb}$ | 100 | 2.54 | 0.083 | 0 | -19.97 | -20.21 |
| | 2017 | $RBM_{TrPrTr}$ | 100 | 2.35 | 0.155 | 31 | -21.52 | -22.05 |
| | 50000 | RBM | 100 | 1.71 | 1 | 0 | -251.44 | -300.89 |
| | 50977 | XBM | 4500 | 2.17 | 0.127 | 0 | -284.59 | **-289.47** |
| | 50609 | $RBM_{FixProb}$ | 4500 | 3.43 | 0.048 | 0 | -226.90 | -293.74 |
| NIPS-0-12 | 43569 | $RBM_{TrPrTr}$ | 4500 | 3.00 | 0.040 | 7 | -309.68 | -525.63 |
| | 5144 | XBM | 100 | 2.22 | 0.113 | 0 | -274.07 | -287.43 |
| | 4966 | $RBM_{FixProb}$ | 100 | 2.74 | 0.078 | 0 | -272.95 | -286.77 |
| | 5220 | $RBM_{TrPrTr}$ | 100 | 2.29 | 0.164 | 41 | -253.09 | -289.62 |
| | 12800 | RBM | 100 | 1.49 | 1 | 0 | -39.40 | -39.58 |
| | 13053 | XBM | 1200 | 2.14 | 0.190 | 0 | -33.07 | **-33.08** |
| | 12957 | $RBM_{FixProb}$ | 1200 | 2.80 | 0.070 | 0 | -40.03 | -40.16 |
| OCR-letters | 14139 | $RBM_{TrPrTr}$ | 1200 | 2.83 | 0.075 | 12 | -44.17 | -45.15 |
| | 1710 | XBM | 100 | 2.36 | 0.154 | 0 | -45.70 | -45.68 |
| | 1743 | $RBM_{FixProb}$ | 100 | 2.49 | 0.083 | 0 | -49.20 | -49.10 |
| | 1960 | $RBM_{TrPrTr}$ | 100 | 2.58 | 0.127 | 4 | -63.72 | -63.69 |
| | 15000 | RBM | 100 | 1.51 | 1 | 0 | -52.04 | -52.50 |
| | 14797 | XBM | 1400 | 2.15 | 0.162 | 0 | -49.22 | **-49.68** |
| | 15003 | $RBM_{FixProb}$ | 1400 | 2.90 | 0.066 | 0 | -50.06 | -50.59 |
| RCV1 | 27555 | $RBM_{TrPrTr}$ | 1400 | 2.58 | 0.081 | 50 | -57.05 | -59.47 |
| | 1992 | XBM | 100 | 2.35 | 0.151 | 0 | -52.15 | -52.30 |
| | 1994 | $RBM_{FixProb}$ | 100 | 2.49 | 0.081 | 0 | -52.01 | -52.17 |
| | 2999 | $RBM_{TrPrTr}$ | 100 | 2.33 | 0.147 | 15 | -54.68 | -54.94 |
| | 30000 | RBM | 100 | 1.62 | 1 | 0 | -35.46 | -35.43 |
| | 29893 | XBM | 2600 | 2.17 | 0.123 | 0 | -30.00 | **-30.62** |
| | 29780 | $RBM_{FixProb}$ | 2600 | 3.20 | 0.052 | 0 | -45.52 | -46.09 |
| Web | 34041 | $RBM_{TrPrTr}$ | 2600 | 3.33 | 0.070 | 50 | -1114.93 | -1118.98 |
| | 3433 | XBM | 100 | 2.28 | 0.149 | 0 | -33.99 | -33.97 |
| | 3333 | $RBM_{FixProb}$ | 100 | 2.62 | 0.076 | 0 | -38.40 | -38.30 |
| | 1302 | $RBM_{TrPrTr}$ | 100 | 2.42 | 0.360 | 2 | -252.90 | -252.88 |

To summarize this set of experiments performed on 10 binary datasets (i.e. MNIST digits, CalTech 101 Silhouettes, and UCI evaluation suite), we report that XBMs outperform all the other models in 16 out of 18 cases considered. In the other two cases, the winner is once $RBM_{FixProb}$, and once $RBM_{TrPrTr}$. Besides that, a very interesting finding is that in all cases $RBM_{TrPrTr}$ models end up having a small-world topology, as reflected by their average shortest path and cluster coefficient. Similarly, $RBM_{FixProb}$ models reach a very small average shortest path (suitable to be qualified as small-worlds), but we can not consider them pure small-worlds as their average cluster coefficient represents the one obtained by random chance. Still, the better overall performance obtained by XBMs may be explain by the fact that its small-world topology is supplemented by its other designed topological features, i.e. scale-free property, the consideration of local neighborhoods and data distribution. As reflected by experiments (including the ones with real-valued data for GXBMs) these features complements each other, while helping XBMs to model well very different data types.

## 5 Conclusion

In this paper we look at the deep learning basic building blocks from a topological perspective, bringing insights from network science. Firstly, we point out that RBMs and GRBMs are small-world bipartite networks. Secondly, by introducing scale-free constraints in RBMs and GRBMs, while still considering some local neighborhoods of visible neurons, and fitting the most connected visible neurons to the most important data features, we propose two novel types of Boltzmann machine models, dubbed complex Boltzmann machine and Gaussian complex Boltzmann machine. Looking at both artificial and real-world datasets (i.e. Geographical Origin of Music, MNIST digits, CalTech 101 Silhouettes, and UCI evaluation suite) we show that XBM and GXBM obtain better performance than other two sparse models (i.e. $RBM_{FixProb}$/$GRBM_{FixProb}$ and $RBM_{TrPrTr}$/$GRBM_{TrPrTr}$) and we illustrate how they outperform even the fully connected RBM and GRBM, respectively:

1. Given the same number of hidden neurons, our proposed models exhibit much faster computational time thanks to a smaller number of parameters which have to be computed (up to a few orders of magnitude smaller than in RBM and GRBM) and comparable reconstruction capabilities.
2. Given the same number of weights, or implicitly a much higher number of hidden neurons for XBM and GXBM, they significantly outperform RBM and GRBM, respectively.

It is worth noting that as the number of neurons increases the order of magnitude between the number of weights in XBM/GXBM and in RBM/GRBM also increases (e.g. one order of magnitude fewer weights in a XBM/GXBM with 100 visible and 100 hidden neurons, and two orders reduction in a XBM/GXBM with 1000 visible and 1000 hidden neurons). This relation will help increasing the typical number of neurons in deep artificial neural networks from the few hundred thousands of today (Krizhevsky et al, 2012; Ba and Caruana, 2014) to even billions in the near-future. In turn this will lead to the ability to tackle problems having much higher dimensional data - something that is today unfeasible without performing dimensionality reduction. For instance, when working on ordinary images[7] today is still a common practice to first extract features using standard image processing techniques, and just those features can be served as inputs to deep models - an example can be found in Srivastava and Salakhutdinov (2012). We speculate that another significant benefit of using sparse topologies, as in XBM/GXBM, would be the ability to better disentangle the features extracted automatically by the hidden layer.

To conclude, in this article, we have shown empirically on 12 datasets that our proposed models, i.e. XBMs and GXBMs, achieve a very good performance as generative models. We mention that more research has to be done in order to understand why their proposed topology (e.g. the scale-free constraints) makes them to perform so well. Further on, we intend to investigate all these directions and to study analytically how the various parameters of the topology-generation algorithm (implicitly the bipartite graph properties) may increase or decrease the generative and discriminative capabilities of XBMs and GXBMs, respectively.

**Acknowledgements** This research has been partly funded by the European Union's Horizon 2020 project INTER-IoT (grant number 687283), and by the NL Enterprise Agency under the TKI SG-BEMS project of Dutch Top Sector.

[7] Please note that a normal RGB image of 1000 by 1000 pixels has 3000000 dimensions.